\documentclass[10pt,twocolumn,letterpaper]{article}

\usepackage{iccv}              
\usepackage{algorithm}
\usepackage{algpseudocode}
\usepackage{colortbl}
\usepackage{xcolor}
\usepackage{graphicx}
\usepackage{caption}
\usepackage[accsupp]{axessibility}  
\newcommand{\mpage}[2]
{
\begin{minipage}{#1\linewidth}\centering
#2
\end{minipage}
}

\definecolor{iccvblue}{rgb}{0.21,0.49,0.74}
\definecolor{myred}{HTML}{FECACA}     
\definecolor{myyellow}{HTML}{FEF9C3}  
\definecolor{myorange}{HTML}{FED7AA}  
\usepackage[pagebackref,breaklinks,colorlinks,allcolors=iccvblue]{hyperref}
\usepackage{colortbl}
\usepackage{booktabs}

\def\paperID{3729} 
\def\confName{ICCV}
\def\confYear{2025}

\title{Bridging Diffusion Models and 3D Representations: \\ A 3D Consistent Super-Resolution Framework}

\author{%
    Yi-Ting Chen\textsuperscript{1} \hspace{0.5cm} Ting-Hsuan Liao\textsuperscript{1} \hspace{0.5cm} Pengsheng Guo\textsuperscript{2} \\ \hspace{0.5cm} Alexander Schwing\textsuperscript{3} \hspace{0.5cm} Jia-Bin Huang\textsuperscript{1}  \\
    \textsuperscript{1}University of Maryland, College Park  \hspace{1cm} \textsuperscript{2} Carnegie Mellon University \\ \hspace{1cm} \textsuperscript{3} University of Illinois Urbana-Champaign \\
}

\begin{document}

\twocolumn[{
\renewcommand\twocolumn[1][]{#1}
\maketitle

\begin{center}
\centering

\includegraphics[width=\linewidth]{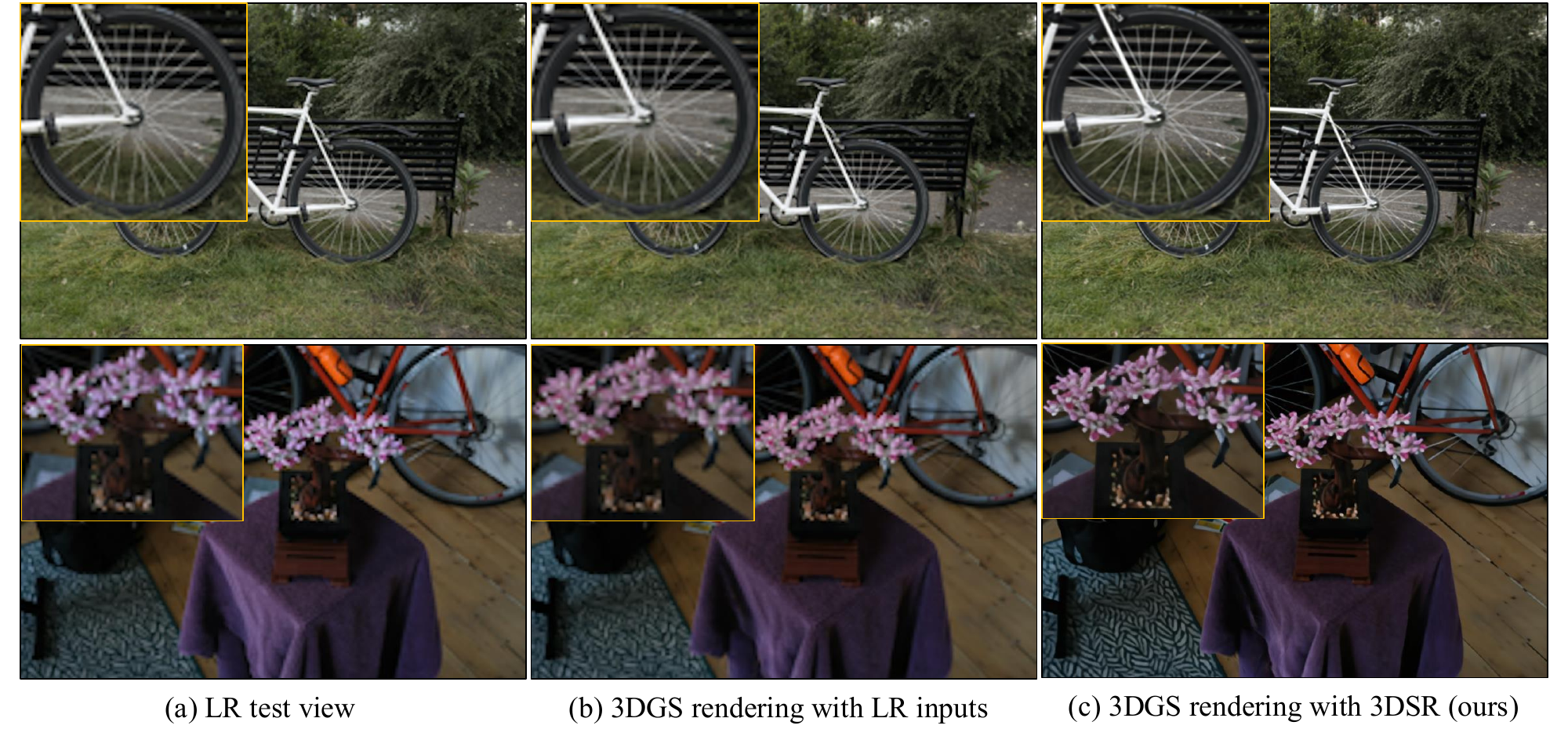}
    \captionof{figure}{Qualitative comparison on the LLFF dataset with a downsampling factor of  $\times8$  and upsampling of  $\times4$. We train 3DGS with low-resolution (LR) inputs and with 3DSR (ours)
    and render test views. The results demonstrate that our method yields fewer artifacts from 3D inconsistencies and better preserves overall structural integrity compared to the baselines.}
    
    \label{fig:teaser}
\end{center}
}]

    

\maketitle
\begin{abstract}

We propose 3D Super Resolution (3DSR), a novel 3D Gaussian-splatting-based super-resolution framework that leverages off-the-shelf diffusion-based 2D super-resolution models. 3DSR encourages 3D consistency across views via the use of an explicit 3D Gaussian-splatting-based scene representation. This makes the proposed 3DSR different from prior work, such as image upsampling or the use of video super-resolution, which either don't consider 3D consistency or aim to incorporate 3D consistency implicitly. Notably, our method enhances visual quality without additional fine-tuning, ensuring spatial coherence within the reconstructed scene. We evaluate 3DSR on MipNeRF360 and LLFF data, demonstrating that it produces high-resolution results that are visually compelling, while maintaining structural consistency in 3D reconstructions. Project page: \url{https://consistent3dsr.github.io/}{}.

\end{abstract}   
\begin{figure*}[t]
\vspace{0.8cm}
    \centering
    \includegraphics[width=\linewidth]{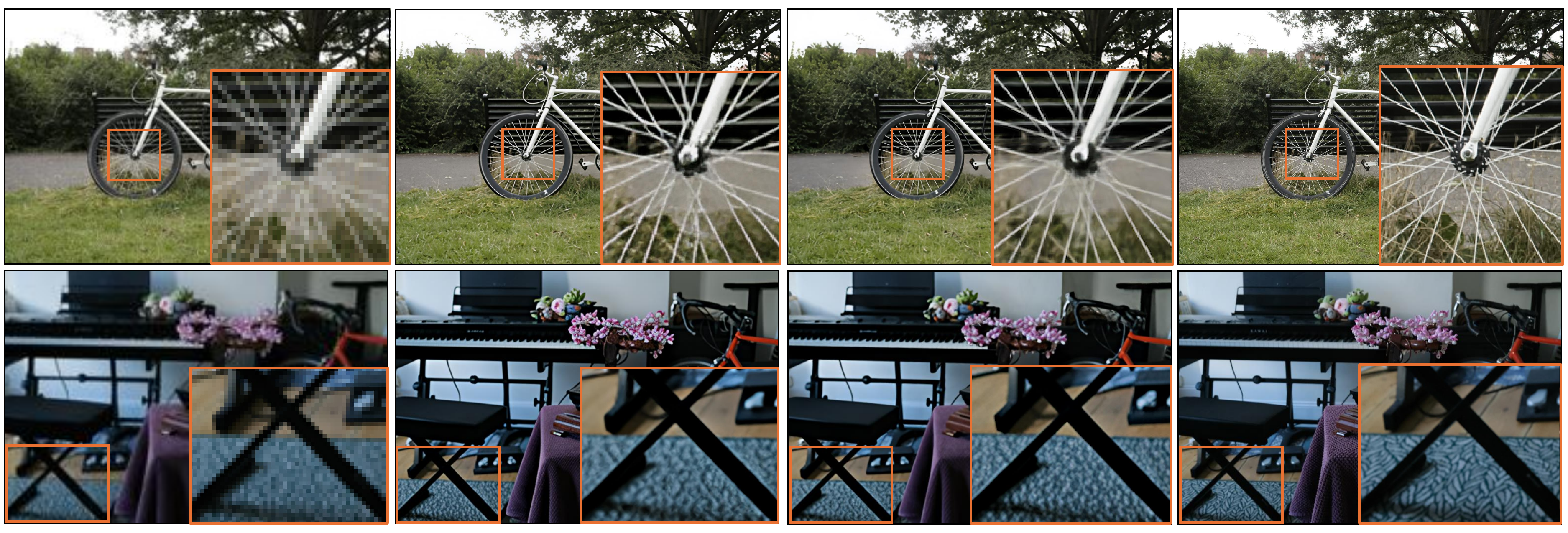}

\mpage{0.24}{{{\small{(a) LR input images}}}}\hfill
\mpage{0.24}{{{\small{(b) StableSR result}}}}\hfill
\mpage{0.24}{{{\small{(c) Rendering of 3DGS trained on StableSR results}}}}
\mpage{0.24}{{{\small{(d) Groundtruth}}}}

    \caption{\textbf{Motivation:} Diffusion-based super-resolution (SR) methods enhance details in high-resolution (HR) images but fail to maintain 3D consistency across views. Given a set of low-resolution (LR) images (a), we apply StableSR~\cite{wang2024exploiting} to generate super-resolved images (b). However, the SR results introduce hallucinated details, as seen in the distorted steel strings of the bike, which appear misaligned and inconsistent. When rendered with 3DGS in (c), these inconsistencies lead to blurring and incorrect geometry. In the second row, the SR output (b) significantly deviates from the ground truth (d), demonstrating the instability of diffusion-based SR. Across multiple views, the inconsistencies in hallucinated textures further degrade 3DGS rendering, resulting in incorrect and incoherent textures.}    
    \label{fig:motivation}
\end{figure*}







\section{Introduction}
\label{sec:intro}






Recent advances in 3D representation learning have significantly improved novel view synthesis (NVS), enabling realistic rendering of scenes from sparse observations. 
Concretely, techniques such as Neural Radiance Fields (NeRFs)~\cite{mildenhall2021nerf} and Gaussian Splatting (GS)~\cite{kerbl3Dgaussians} have demonstrated remarkable success in reconstructing high-fidelity 3D representations, benefiting applications in virtual reality, augmented reality, robotics, and content creation. 
However, despite their effectiveness, these methods are inherently limited by the spatial resolution of the input camera views, resulting in synthesized images that often lack fine details. 
This limitation restricts their applicability in high-quality rendering and immersive real-world experiences.

To address this resolution constraint, a straightforward approach is to upsample low-resolution (LR) training data using either image super-resolution (ISR)~\cite{lim2017enhanced,zhang2018image,liang2021swinir,zamir2022restormer,wang2018esrgan,wang2021towards,wang2024sinsr,wang2024exploiting} or video super-resolution (VSR)~\cite{wang2019edvr, liu2022learning, chan2022basicvsr++, zhou2024upscaleavideo, xu2024videogigagan} techniques. 
However, applying ISR methods independently across views often introduces inconsistencies, leading to artifacts and a loss of geometric coherence. 
In contrast, VSR methods, while designed to enforce temporal consistency explicitly and 3D consistency implicitly, tend to struggle to preserve fine-grained details, resulting in blurred textures compared to ISR techniques. 
An alternative solution is to refine the underlying 3D representation directly, enhancing scene details at the feature level. 
However, existing approaches~\cite{shen2024supergaussian} often require significant computational resources or introduce new artifacts due to the inaccurate reconstruction of high-frequency details.

In parallel, diffusion models~\cite{ho2020denoising,song2021denoising,SongICLR2021,liu2023flow,LipmanICLR2023,albergo2023building,albergo2023stochastic} have recently demonstrated remarkable results, often topping benchmarks in image synthesis, editing, and super-resolution. 
Unlike traditional deep learning models, diffusion-based approaches iteratively refine noisy inputs, enabling the generation of high-frequency textures and realistic details that surpass conventional GAN- or CNN-based super-resolution methods. 
Diffusion-based image super-resolution~\cite{wang2024sinsr,wang2024exploiting} has significantly improved production of photo-realistic details with sharper textures. 
However, these fine details are often \emph{hallucinated} rather than faithfully reconstructed, leading to cross-view inconsistency when applied to multi-view image generation. 
Since diffusion models lack an inherent understanding of 3D geometry, applying them independently to different viewpoints results in inconsistent textures and structural artifacts across synthesized views, akin to classic ISR.

In this work, we leverage the power of diffusion models to address the resolution constraint of the Gaussian-splatting 3D scene representation. 
Specifically, we exploit a diffusion model for high-quality detail generation while encouraging multi-view consistency in novel view synthesis via an integrated Gaussian-splatting representation. 
This approach enhances high-resolution rendering while maintaining structural coherence across multiple viewpoints, effectively addressing a limitation of existing super-resolution and novel view synthesis techniques.

To evaluate our approach, we conduct extensive experiments on two real-world datasets, MipNeRF360~\cite{barron2022mip} and LLFF~\cite{mildenhall2019llff}, and assess performance using PSNR, SSIM, LPIPS~\cite{zhang2018unreasonable}, NIQE~\cite{mittal2012making}, MEt3R~\cite{asim2025met3r}, and FID~\cite{heusel2017gans}. 
As shown in Figure~\ref{fig:teaser} and Figure~\ref{fig:motivation}, experimental results demonstrate that the proposed method improves perceptual quality, geometric consistency, and rendering fidelity compared to existing state-of-the-art techniques.

\section{Related Work}
\label{sec:related}

\subsection{3D Representations}
Recent advances in 3D representations have significantly improved novel view synthesis (NVS), enabling high-quality scene rendering from sparse inputs. Neural Radiance Fields (NeRFs)~\cite{mildenhall2021nerf} model scenes as continuous volumetric functions, achieving photo-realistic synthesis but requiring dense input views and long training times. 
Variants such as Mip-NeRF~\cite{barron2021mip} and Instant-NGP~\cite{mueller2022instant} improve anti-aliasing and training efficiency, yet resolution constraints remain a challenge, leading to blurry or low-detail renderings at high resolutions.

3D Gaussian Splatting (3DGS)~\cite{kerbl3Dgaussians} offers a more efficient alternative, representing scenes as adaptive 3D Gaussians, enabling real-time rendering without ray marching. 
Mip-Splatting~\cite{yu2024mip} further enhances 3DGS by integrating multi-scale representations, reducing aliasing and improving view extrapolation. 

While our method is compatible with multiple 3D representations, we adopt 3DGS 
due to its better training efficiency compared to NeRF-based approaches. 
This allows us to focus on improving high-resolution rendering and 3D consistency without the computational overhead of traditional volume-based representations.

\subsection{Super-Resolution}

Super-resolution plays a crucial role in enhancing low-resolution (LR) images and videos, with applications in computational photography, graphics, and 3D scene reconstruction. 
Recent advances in deep learning-based SR have significantly improved reconstruction quality, with approaches focusing on image super-resolution  and video super-resolution. While discriminative models and generative adversarial networks (GANs) have been widely explored for super-resolution, diffusion-based approaches have recently emerged as a powerful alternative, offering improved texture reconstruction and perceptual quality.  

\noindent\textbf{Image Super-Resolution (ISR).} 
ISR aims to reconstruct a high-resolution (HR) image from a single LR input. Traditional CNN-based ISR methods, such as EDSR~\cite{lim2017enhanced} and RCAN~\cite{zhang2018image}, leverage residual learning and channel attention to recover fine-grained textures. Transformer-based SR models, including SwinIR~\cite{liang2021swinir} and Restormer~\cite{zamir2022restormer}, have further advanced the field by utilizing self-attention mechanisms to improve feature aggregation and enhance reconstruction quality.
Generative models, such as ESRGAN~\cite{wang2018esrgan} and GFP-GAN~\cite{wang2021towards}, employ generative adversarial networks (GANs) to address the inherently ambiguous nature of the super-resolution problem. To synthesize photo-realistic details, GAN-based methods hallucinate fine textures and improve perceptual quality. However, these models are typically focused on specific categories, such as faces~\cite{wang2021towards}, and may struggle to generalize across more complex real-world SR tasks. 
More recently, diffusion models have emerged as a promising alternative, surpassing traditional discriminative and GAN-based methods in ISR. Instead of directly learning a mapping between LR and HR images, diffusion-based SR models iteratively refine images through a denoising process. Models like SinSR~\cite{wang2024sinsr} and StableSR~\cite{wang2024exploiting} excel in producing highly detailed and texture-rich images in large-scale real world datasets. However, a key limitation of diffusion-based ISR methods is their lack of geometric awareness, which can lead to challenges in maintaining structural consistency, particularly when applied to 3D scene synthesis.

\noindent\textbf{Video Super-Resolution (VSR).}
In VSR, the goal is to enhance video sequences while preserving temporal consistency. EDVR~\cite{wang2019edvr} and BasicVSR++~\cite{chan2022basicvsr++} improve spatio-temporal feature aggregation, effectively reducing flickering artifacts in enhanced frames. Recent transformer-based methods, such as TTVSR~\cite{liu2022learning}, integrate spatial and temporal self-attention to achieve strong performance in enhancing video resolution. Recently, VideoGigaGAN~\cite{xu2024videogigagan} has emerged as a promising method, employing a generative adversarial network (GAN) to upscale videos focusing on both perceptual quality and temporal consistency. This approach has demonstrated impressive results in preserving high-frequency details while maintaining coherence across multiple frames. Additionally, Upscale-A-Video~\cite{zhou2024upscaleavideo}, another recent method, adopts a text-based diffusion model and improves temporal coherence through local temporal layers and a global recurrent latent propagation module.
Despite these advances, VSR methods often enforce temporal consistency via implicit spatio-temporal aggregation mechanisms~\cite{wang2019edvr, liu2022learning, chan2022basicvsr++, zhou2024upscaleavideo} or flow-guided propagation~\cite{xu2024videogigagan, zhou2024upscaleavideo}, but they do not guarantee 3D view consistency. This lack of geometric awareness can lead to inconsistencies in spatial coherence when applied to multi-view scene reconstruction. Furthermore, VSR techniques frequently require multi-frame alignment, which increases computational complexity, particularly when handling high-resolution videos or large-scale 3D scene synthesis. These challenges can result in jittery, less detailed reconstructions and higher computational costs.

\noindent\textbf{Super-Resolution in 3D.}
Recent works have explored enhancing the quality of 3D assets using super-resolution techniques, but naively applying ISR methods to individual frames often leads to 3D inconsistencies, as these methods treat each view independently, disregarding geometric coherence. To maintain consistency with the LR input, some methods incorporate subpixel loss functions that align super-resolved outputs with the original LR images, such as NeRF-SR~\cite{wang2022nerf} and SRGS~\cite{feng2024srgs}, which preserve 2D details but fail to ensure cross-view consistency, potentially causing structural distortions in novel views. To address multi-view consistency, some approaches leverage temporal coherence from VSR, like SuperGaussian~\cite{shen2024supergaussian}, which extends VSR principles to 3D Gaussian Splatting but requires fine-tuning the video model, limiting its general applicability. It relies on frame-based SR techniques and is constrained by the temporal limitations of video models, which don't explicitly address 3D scene consistency. Additionally, some methods enhance model representation when high-resolution input data is available, such as 4K-NeRF~\cite{wang20224k} and UHD-NeRF~\cite{li2023uhdnerf}. This improves rendering quality with high-resolution NeRF architectures. Further,  SRGS~\cite{feng2024srgs} and GaussianSR~\cite{hu2024gaussiansr, yu2024gaussiansr} focus on generating high-resolution details for 3D Gaussians. But these methods improve the 3D representation rather than ensuring 3D consistency, which distinguishes them from our goal of upsampling a 3D scene in a 3D-consistent manner.

Our approach differs from existing methods by combining diffusion-based super-resolution with a 3D representation, ensuring consistency at the 3D scene level instead of relying on frame-by-frame super-resolution. We build upon diffusion-based super-resolution while addressing multi-view consistency in 3D scene synthesis. By integrating diffusion priors into a view-consistent enhancement module, our method generates high-resolution outputs and preserves spatial coherence across multiple viewpoints, all without needing to fine-tune VSR models or rely on high-resolution input.


\subsection{View Consistent Diffusion Model}


Recent advances in diffusion models have enabled impressive image, video, and 3D content generation, but maintaining consistency across multiple views remains a major challenge. Existing approaches to consistent diffusion-based generation can be broadly categorized into 3D object-level methods and 2D multi-view consistency techniques.

\noindent\textbf{3D Object-Level Consistency.} 
Several recent works have tackled 3D-consistent object synthesis using diffusion models. SyncDreamer~\cite{liu2023syncdreamer} and MVDream~\cite{shi2023mvdream} extend text-to-image diffusion models by enforcing multi-view consistency at the object level, enabling high-quality, consistent 3D object generation from a few input views. These methods typically leverage image-space priors and cross-view attention mechanisms to generate multiple perspectives of an object, while preserving structural consistency. However, they primarily focus on isolated objects rather than entire scenes. As a result, they struggle to model complex spatial relationships between multiple elements in a 3D environment, limiting their applicability to full-scene generation.

\noindent\textbf{2D Multi-View Consistency.} 
Other approaches emphasize enforcing consistency at the 2D level by refining multi-view diffusion outputs. MultiDiffusion~\cite{bar2023multidiffusion} improves spatial consistency in image generation by synchronizing the denoising process across overlapping regions of multiple views. Generative Power of Ten~\cite{wang2024generative} extends this concept by incorporating a joint multi-scale sampling approach, averaging noise across multiple views from different scales to encourage alignment. While these techniques improve 2D consistency, they cannot be directly applied to 3D scenes due to the lack of explicit geometry modeling, which results in missing feature correspondences across different views. 

In this work, we build upon the idea of noise averaging for consistency but extend it to the 3D scene level. Instead of enforcing consistency purely in 2D space, we leverage a 3D representation to directly regulate multi-view coherence. By integrating diffusion priors with 3D structural constraints, our method enhances high-resolution details while preserving geometric fidelity across all synthesized views. This enables full-scene super-resolution without the inconsistency issues inherent in 2D-only averaging techniques.

\begin{figure*}[t]
    \centering
    \includegraphics[width=\linewidth]{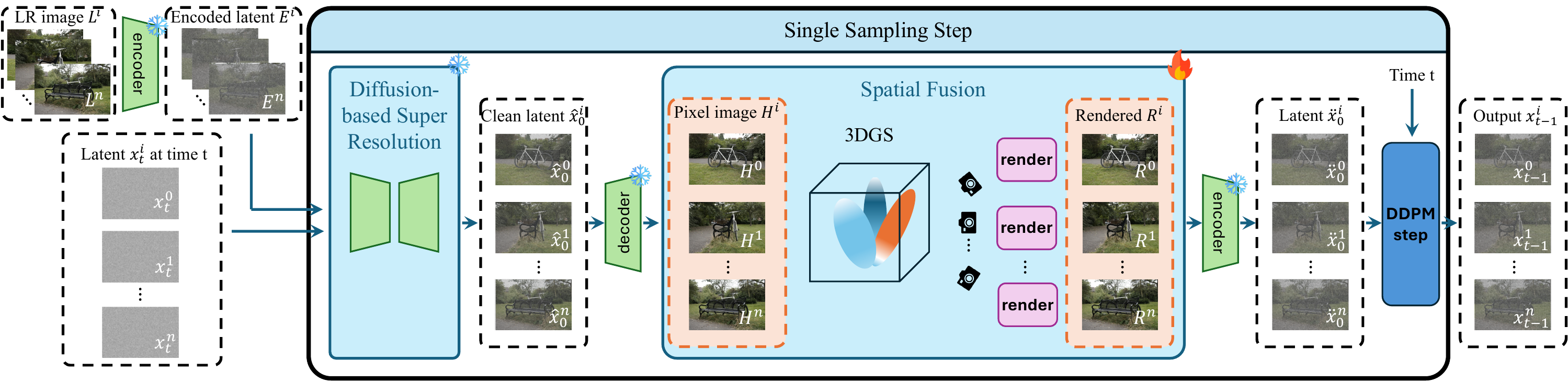}
    \caption{\textbf{Overview of one single sampling step.}
(a) The diffusion-based super-resolution method takes the latent representation  $x_t^i$ of $i$th image at time step  $t$  and the encoded low-resolution (LR) image latent  $E^i$  as inputs to predict the clean latent  $\hat{x}_0^i$
(b) The predicted clean latent  $\hat{x}_0^i$  is then decoded into a high-resolution (HR) image  $H^i$ , constrained by the LR image latents to ensure consistency.
(c) The super-resolved images $H^i$  are subsequently used as inputs for a 3D Gaussian Splatting model pretrained on LR images. By leveraging the 3D representation, the rendered images  $R^i$  are encouraged to exhibit improved 3D consistency, facilitating spatial fusion across different views.
(d) After obtaining the rendered images  $R^i$  from the 3D Gaussian Splatting model trained on SR images, they are encoded into the latent space as a 3D-consistent clean latent  $\ddot{x}_0^i$
(e) The 3D-consistent latent  $\ddot{x}_0^i$  is then input into the diffusion model along with the latent at time step $t$ ,  $x_t^i$ , to perform a denoising step, yielding the updated latent  $x_{t-1}^i$.}
    \label{fig:overview}
\end{figure*}

\section{Method}\label{sec:method}
We propose 3DSR, a 3D Gaussian-splatting-based super-resolution framework that leverages diffusion-based 2D super-resolution to enhance 3D consistency across different views. Note, the diffusion-based SR method used in this framework is not constrained to single image super-resolution (ISR)  or video super-resolution (VSR), enabling 3DSR to benefit from future advances in this area. 
Compared to VSR-based methods like SuperGaussian~\cite{shen2024supergaussian}, our approach encourages 3D consistency more explicitly without fine-tuning the video model.

In this section, we will first present preliminaries of diffusion models in Section~\ref{method:prelim}. We will then introduce the problem definition and our proposed multiview consistent noise sampling in Section~\ref{method:problem_definition}, and the losses we use to optimize the 3D Gaussians in Section~\ref{method:optimization}. 


\subsection{Preliminaries} \label{method:prelim}



\noindent\textbf{Diffusion Models.} 
Diffusion models iteratively refine an image by reversing a noise perturbation process. The forward diffusion process transforms a clean image $\mathbf{x}_0$ into a noisy latent representation $\mathbf{x}_t$ over $t$ steps by adding Gaussian noise:
\begin{equation}
q(\mathbf{x}_t \mid \mathbf{x}_0) = \mathcal{N}(\mathbf{x}_t; \sqrt{\bar{\alpha}_t} \mathbf{x}_0, (1 - \bar{\alpha}_t) \mathbf{I}).
\label{eq:forward_process}
\end{equation}
Here $\bar{\alpha}_t = \prod_{s=1}^{t} \alpha_s$ is the cumulative noise schedule, ensuring a smooth transition between clean and noisy images. Based on Equation~\ref{eq:forward_process}, the explicit expression of the noisy image at time step $t$ reads as follows:
\begin{equation}
\mathbf{x}_t = \sqrt{\bar{\alpha}_t} \mathbf{x}_0 + \sqrt{1 - \bar{\alpha}_t} \boldsymbol{\epsilon}, \quad \boldsymbol{\epsilon} \sim \mathcal{N}(\mathbf{0}, \mathbf{I}).
\label{eq:reparam}
\end{equation}
This formulation is useful for training, as it allows the model to predict the noise $\boldsymbol{\epsilon}$ directly. 

To reverse the diffusion process, a neural network $\epsilon_\theta(\mathbf{x}_t, t)$ is trained to estimate $\boldsymbol{\epsilon}$, enabling denoising at each step. The posterior distribution for sampling the previous step $\mathbf{x}_{t-1}$ is
\begin{equation}
p_\theta(\mathbf{x}_{t-1} \mid \mathbf{x}_t) = \mathcal{N}(\mathbf{x}_{t-1}; \mu_\theta(\mathbf{x}_t, t), \sigma_t^2 \mathbf{I}),
\label{eq:reverse_process}
\end{equation}
where the mean is computed as follows:
\begin{equation}
\mu_\theta(\mathbf{x}_t, t) = \frac{1}{\sqrt{\alpha_t}} \left(\mathbf{x}_t - (1 - \alpha_t) \epsilon_\theta(\mathbf{x}_t, t) \right).
\label{eq:mean_estimation}
\end{equation}
The updated latent variable is then obtained using
\begin{equation}
\mathbf{x}_{t-1} = \mu_\theta(\mathbf{x}_t, t) + \sigma_t \mathbf{z}, \quad \mathbf{z} \sim \mathcal{N}(\mathbf{0}, \mathbf{I}).
\label{eq:denoising_step}
\end{equation}

Instead of reconstructing $\mathbf{x}_0$ step by step through multiple denoising iterations, it can be directly estimated using:
\begin{equation}
\hat{\mathbf{x}}_0 = \frac{1}{\sqrt{\bar{\alpha}_t}} \left(\mathbf{x}_t - \sqrt{1 - \bar{\alpha}_t} \epsilon_\theta(\mathbf{x}_t, t) \right).
\label{eq:clean_latent}
\end{equation}
This formulation allows the model to recover a clean signal without iterating through all time steps. Note that the relationship between $\mathbf{x}_{t-1}$  and  $\mathbf{x}_0$ can be explicitly expressed by substituting  $\mathbf{x}_0$  from Equation~\eqref{eq:clean_latent} into Equation~\eqref{eq:denoising_step}, resulting in:
\begin{equation}
\mathbf{x}_{t-1} = \sqrt{\alpha_{t-1}} \hat{\mathbf{x}}_0 + \eta_t \epsilon_\theta(\mathbf{x}_t, t) + \sigma_t \epsilon_t,
\label{eq:simplified_xt-1}
\end{equation}
where $\eta_t$ = $\sqrt{1 - \alpha_{t-1} - \sigma_t^2}$ determines the contribution of the noise component to the update step.


\subsection{Proposed Framework} \label{method:problem_definition}
The overall framework of our proposed method is illustrated in Figure~\ref{fig:overview}. In our approach, the encoder, decoder, and diffusion-based model remain frozen, while only the 3D representation is optimized. Notably, the 3D representation is flexible and can be instantiated as a NeRF, Gaussian Splatting, or any alternative representation that effectively encodes 3D scene information.

Given a set of low-resolution (LR) input images $\{L^i\}_{i=0}^{n}$ and corresponding camera poses $\{P^i\}_{i=0}^n$, our objective is to reconstruct a high-quality 3D scene. These LR images serve as inputs for training the 3D representation  $\Theta$, yielding an LR-pretrained model. Additionally, the LR images are encoded by the encoder $ \mathcal{E}(\cdot) $ into latent space representations  $E^i$, which act as a conditioning signal for the diffusion-based super-resolution model as well as the decoder $ \mathcal{D}(\cdot)$. This effectively encourages that the output remains faithful to the original LR observations.

At the initial sampling step  $t$, a set of randomly sampled noise latents  $x_t^i$  is initialized as the input to the diffusion-based SR model, alongside the LR image latents  $E^i$. The model then estimates the clean image latent  $\hat{x}_0^i$  using Equation~\eqref{eq:clean_latent}. This estimated latent is subsequently decoded into image space, producing an intermediate high-resolution image  $H^i$.

To encourage 3D consistency across different views, we leverage the ability of  $\Theta$  to generate spatially coherent representations. The estimated high-resolution images  $H^i$  are used as training data for the 3D scene representation $\Theta$, allowing it to render the corresponding high-resolution images with given pose $\Theta(P^i)=R^i$  from the same viewpoints. This process regularizes the super-resolution output, ensuring that fine details are structurally aligned across multiple views.

The 3D-consistent high-resolution rendering  $R^i$  is then re-encoded into the latent space. We use  $\ddot{x}_0^i$ to refer to this latent space encoding. Compared to the initial estimate  $\hat{x}_0^i$, the updated latent  $\ddot{x}_0^i$  better preserves multi-view coherence. Using Equation~\eqref{eq:denoising_step}, we incorporate the updated 3D-consistent latent  $\ddot{x}_0^i$  along with the current latent  $x_t^i$, enabling us to refine the denoising process and obtain the improved latent $x_{t-1}^i$ for the next iteration. The entire algorithm is summarized in Algorithm~\ref{alg:3d_diffusion_sr}.



\begin{algorithm}[t]
\caption{3D-Consistent Diffusion-Based Super-Resolution}
\label{alg:3d_diffusion_sr}
\begin{algorithmic}[1] 

\State \textbf{Input:} LR images  $\{L^i\}_{i=0}^{n}$, camera poses $\{P^i\}_{i=0}^n$, encoder $\mathcal{E}$, decoder $\mathcal{D}$, super resolution diffusion model $\epsilon_\theta$
\State \textbf{Initialization:} pretrained 3D representation $\Theta_{LR}$, $\{x_t^i\}_{i=0}^{n} \sim \mathcal{N}(\mathbf{0}, \mathbf{I})$, LR latents $E^i =  \mathcal{E}(L^i)$

\For{$t = T$ to $0$} 
    \For{$i = 0$ to $n$}
        \begin{equation}
        \hat{x}_0^i = \frac{1}{\sqrt{\bar{\alpha}_t}} \left(x_t^i - \sqrt{1 - \bar{\alpha}_t} \epsilon_\theta(x_t^i, t, E^i) \right)
        \end{equation}
        \State $H^i =  \mathcal{D}(\hat{x}_0^i, E^i)$

    \EndFor
    \State Train $\Theta$ with $\{H^i\}_{i=0}^n$
    \State $\{R^i\}_{i=0}^n$ = render($\Theta, \{P^i\}_{i=0}^n)$
    \For{$i = 0$ to $n$}
        \State $\ddot{x}_0^i$ =  $\mathcal{E}(R^i)$
        \State Denoise one step
        \begin{equation}
        x_{t-1}^i = \sqrt{\alpha_{t-1}} \ddot{x}_0^i + \eta_t \epsilon_\theta(x_t^i, t) + \sigma_t \epsilon_t
        \end{equation}
    \EndFor
\EndFor

\State \textbf{Output:} Optimized 3D representation $\Theta^*$ with high-resolution consistency.
\end{algorithmic}
\end{algorithm}

\subsection{Training Objective} \label{method:optimization}
To optimize 3D Gaussian Splatting (3DGS) for higher fidelity, we adopt the subsampling-based regularization strategy proposed in~\cite{wang2022nerf, feng2024srgs}. This approach mitigates inconsistencies in high-resolution details by subsampling the rendered image  $R^i$  to match the resolution of the low-resolution input  $L^i$, producing  $R_\text{lr}^i$. This encourages alignment between the rendered and the input images, improving overall consistency during rendering optimization.

Our overall loss function,  $\mathcal{L}_\text{all}$, consists of two terms:
\begin{equation}
\mathcal{L}_\text{all} = \mathcal{L}_\text{hr}(H^i, R^i) + \lambda\mathcal{L}_\text{lr}(L^i, R_{lr}^i).
\end{equation}
Here,  $\lambda$ is a weighting factor controlling the contribution of the low-resolution subsampling loss $\mathcal{L}_\text{lr}$.

For both  $\mathcal{L}_\text{hr}$  and  $\mathcal{L}_\text{lr}$, we follow Mip-Splatting~\cite{yu2024mip}, employing a combination of  an $\mathcal{L}_1$  loss and a D-SSIM loss  $\mathcal{L}_\text{D-SSIM}$, i.e.,
\begin{equation}
\mathcal{L}_\alpha = (1-\delta)\mathcal{L}_1^\alpha + \delta\mathcal{L}_\text{D-SSIM}^\alpha,
\end{equation}
where $\alpha\in\{\text{hr},\text{lr}\}$. 
Further,  $\delta$  is a hyperparameter that balances the contribution of the $\mathcal{L}_1$ loss and the  $\mathcal{L}_\text{D-SSIM}$ loss.
\section{Experiments}
\label{sec:exp}
\subsection{Experimental Setup}
\noindent\textbf{Dataset.} 
We evaluate our approach on two real-world datasets, MipNeRF360~\cite{barron2022mip} and LLFF~\cite{mildenhall2019llff}. 
The MipNeRF360 dataset consists of high-resolution images at its native $\times1$ scale (approximately $3000 \times 4000$ pixels) and includes both indoor and outdoor scenes. To simulate different resolution scenarios, we apply two downsampling factors: $\times16$ with subsequent upsampling to $\times4$, and $\times8$ with upsampling to $\times2$. 
This use of two downsampling factors encompasses two distinct scenarios: one in which the input consists of low-resolution images, and another where moderately high-resolution images are used for upsampling. 

In contrast, the LLFF dataset contains images captured from forward-facing camera poses and provides significantly fewer frames for both training and testing (approximately $20$–$40$) compared to MipNeRF360, which provides around $150$–$200$ frames. We adhere to the testing protocol of DiSR-NeRF~\cite{lee2024disr}, where images are downsampled by a factor of $\times8$ and subsequently upsampled to $\times2$.

To assess rendering quality, for both datasets we follow the protocol used by prior work~\cite{mildenhall2021nerf, kerbl3Dgaussians}, removing every $8$th frame from the training set and designating it as a test image.


\begin{figure*}[t]
    \centering
    \includegraphics[width=\linewidth]{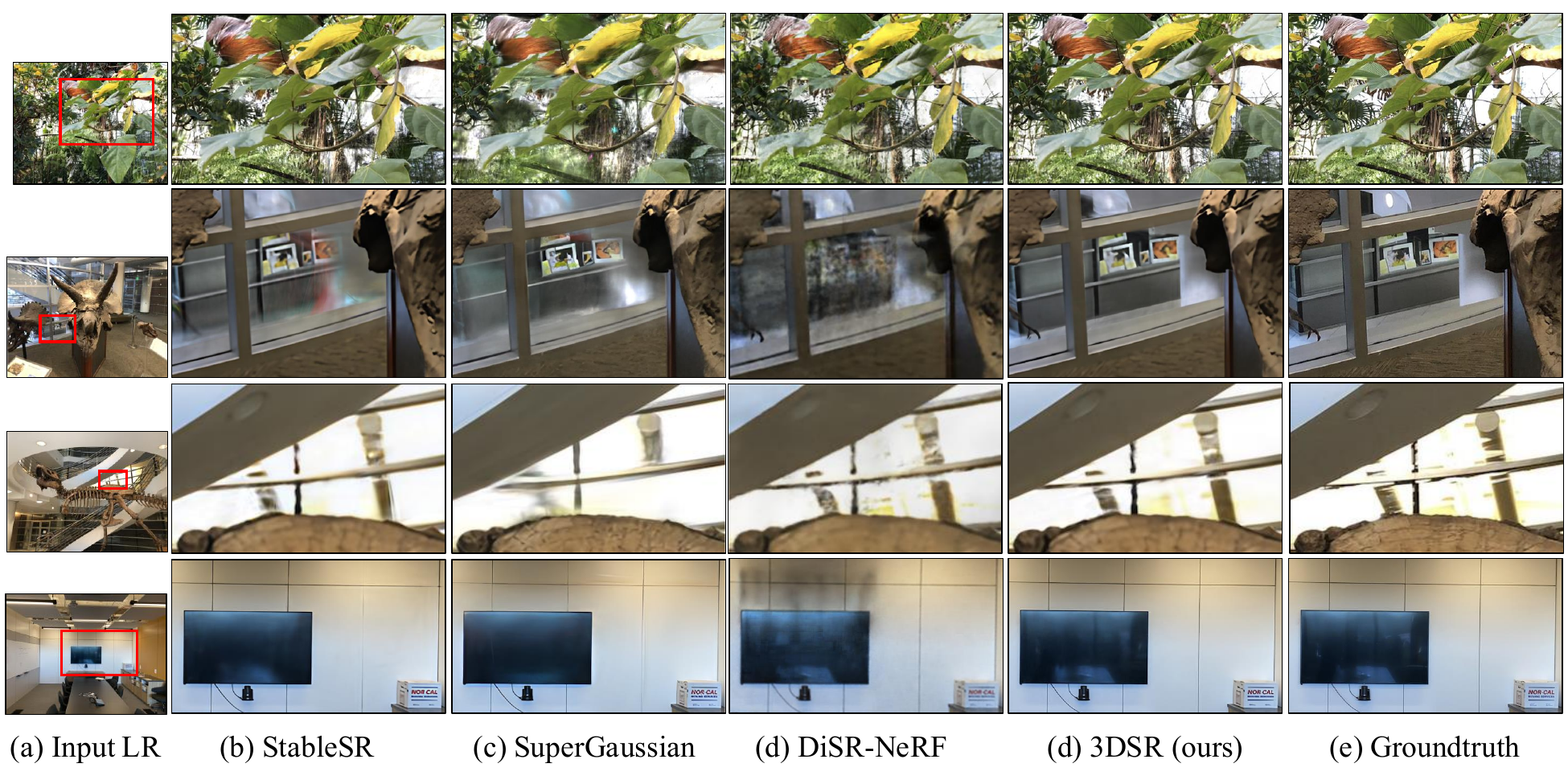}
    \caption{Qualitative comparison on the LLFF dataset with a downsampling factor of  $\times8$  and upsampling of  $\times4$. We train 3DGS using different super-resolved (SR) images and render novel views. The results demonstrate that our method yields fewer artifacts from 3D inconsistencies and better preserves overall structural integrity compared to the baselines.}
    
    \label{fig:qual_llff_ds_8}
\end{figure*}

\begin{figure*}[t]
\vspace{0.5cm}
    \centering
    \includegraphics[width=\linewidth]{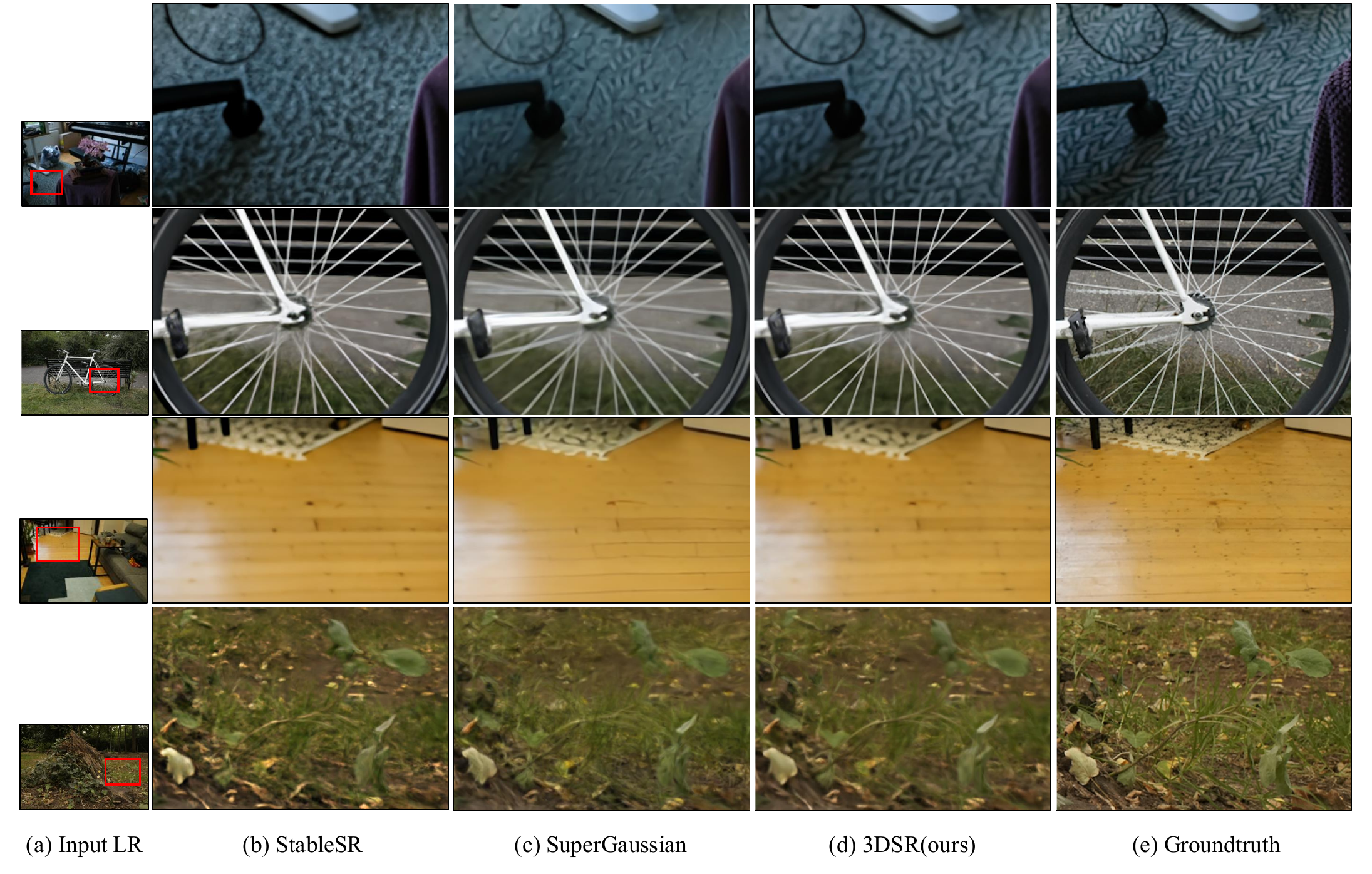}
    \caption{Qualitative comparison on the MipNeRF360 dataset with a downsampling factor of  $\times16$  and upsampling of  $\times4$. We train 3DGS using different super-resolved (SR) images and render novel views. To demonstrate the effectiveness of our method across diverse real-world scenarios, we present results on both indoor and outdoor scenes. Our results demonstrate that our approach preserves structural integrity and fine textures better than the baselines.}
    
    \label{fig:qual_mip_ds_16}
\end{figure*}

\noindent\textbf{Metrics.} 
We evaluate our method using traditional metrics such as PSNR, SSIM, and LPIPS~\cite{zhang2018unreasonable}, the no-reference image quality metric NIQE~\cite{mittal2012making}, the 3D consistency metric MEt3R~\cite{asim2025met3r}, and the FID~\cite{heusel2017gans}, which measures the distributional similarity between generated and ground-truth images in feature space.

\noindent\textbf{Baselines.} 
We compare our approach with three representative baselines: StableSR~\cite{wang2024exploiting}, a diffusion-based single-image super-resolution (ISR) method; DiSR-NeRF~\cite{lee2024disr}, a diffusion-based 3D super-resolution method; and SuperGaussian~\cite{shen2024supergaussian}, a video super-resolution (VSR)-based approach for 3D super-resolution.

StableSR serves as a baseline to evaluate 3D consistency in single-image-based SR. Since each image is super-resolved independently, it does not account for inter-view coherence, often leading to 3D inconsistencies. To evaluate StableSR, we super-resolve each low-resolution (LR) image independently and feed the resulting high-resolution (HR) images into Mip-Splatting~\cite{yu2024mip} for 3D reconstruction.

DiSR-NeRF is more closely related to our method, as it also integrates diffusion models into 3D super-resolution. However, it relies on Score Distillation Sampling and ancestral sampling to optimize a NeRF representation through the supervision of latent noise predictions, which is different from our $x_0$ guided 3DGS-based strategy. Note that DiSR-NeRF is not compatible with the MipNeRF360 dataset, so we only include comparisons on the LLFF dataset.

SuperGaussian, uses a VSR-based pipeline, leveraging video models to promote 3D consistency. It enhances a sequence of LR images via a video SR model before 3D reconstruction. Since our datasets are not temporally consistent, we synthesize a continuous video sequence by interpolating three intermediate frames between each pair of training views. We render a low-resolution video along this trajectory and apply a VSR model~\cite{chan2022investigating}, then use its output frames to train a Gaussian Splatting model~\cite{kerbl20233d}.

\noindent\textbf{Implementation details.} 
In our experiments, we adopt Mip-Splatting~\cite{yu2024mip} as 3D representation and follow its original setup, training for 30,000 iterations using LR images. For our method, we use StableSR-Turbo~\cite{wang2024exploiting}, a diffusion-based SR model trained to produce high-quality outputs in just four denoising steps, significantly improving runtime efficiency without compromising quality. We set the diffusion sampling steps to 4, with 5,000 3DGS training iterations per step. During each step, we set the subsampling weight to 1, $\lambda = 1$, and $\delta = 0.2$. All experiments are conducted on an Nvidia A6000 GPU with 49GB of memory.

\subsection{Results and Analysis}
\noindent\textbf{Quantitative results.}
We evaluate our method against baselines including SuperGaussian and StableSR on both the LLFF and MipNeRF360 datasets, and against DiSR-NeRF on LLFF only, due to code incompatibility with MipNeRF360. As shown in Table~\ref{tab:LLFF_quant} and Table~\ref{tab:mipnerf360_ds_8}, 3DSR(ours) outperforms all baselines across most evaluation metrics. The traditional metrics PSNR, SSIM, and LPIPS are computed with respect to reference ground truth images, indicating that our approach produces higher-fidelity renderings. NIQE evaluates the perceptual quality of a single image, where our method ranks second best. MEt3R, which measures multiview 3D consistency, shows a clear improvement from the single-image baseline StableSR to our method, confirming the enhanced geometric consistency. In terms of FID, our method produces outputs whose distribution is closest to that of the ground truth images.

\noindent\textbf{Qualitative results.} 
%
The qualitative results are presented in Figure~\ref{fig:qual_llff_ds_8} and Figure~\ref{fig:qual_mip_ds_16}. In Figure~\ref{fig:qual_llff_ds_8}, SuperGaussian yields blurrier outputs, likely due to the limited number of training views in LLFF, which hampers its ability to enforce cross-view consistency. StableSR introduces artifacts and geometric distortions, as its hallucinated high-frequency details do not align well across views, resulting in blurring in high-texture regions (e.g., row 1). DiSR-NeRF often produces black and blurry artifacts, further degrading visual quality. In contrast, 3DSR preserves structural details in both complex textures (row 1) and simpler features such as lines and text (row 4). 

In Figure~\ref{fig:qual_mip_ds_16}, SuperGaussian and StableSR produce misaligned and blurry novel views, struggling to maintain fine structural fidelity. Our method demonstrates stronger geometry preservation and significantly reduces 3D inconsistency artifacts.

These results demonstrate that our method better preserves geometry and yields fewer artifacts than baseline methods, reinforcing the effectiveness of our diffusion-based 3D super-resolution, which encourages consistency. 

\begin{table}[t]
\centering
\scriptsize
\caption{LLFF $\times 8$ downsampled and $\times4$ upsampled.}
\vspace{-3.5mm}
\resizebox{\columnwidth}{!}{
\begin{tabular}{l|cccccc}
\toprule
\textbf{Method} & \textbf{PSNR}$\uparrow$ & \textbf{SSIM}$\uparrow$ & \textbf{LPIPS}$\downarrow$ & \textbf{NIQE}$\downarrow$ & \textbf{MEt3R}$\downarrow$ & \textbf{FID}$\downarrow$ \\
\midrule
SuperGaussian~\cite{shen2024supergaussian} & \cellcolor{myorange}23.054 & \cellcolor{myorange}0.725 & \cellcolor{myyellow}0.296 & \cellcolor{myred}4.553 & 0.541 & \cellcolor{myyellow}51.199 \\
DiSR-NeRF~\cite{lee2024disr}    & 22.504     & 0.697    & 0.310    & 6.293    & \cellcolor{myorange}0.518 & 54.138     \\
StableSR~\cite{wang2024exploiting}      & \cellcolor{myyellow}22.748 & \cellcolor{myyellow}0.717 & \cellcolor{myorange}0.219 & \cellcolor{myyellow}4.793 & \cellcolor{myyellow}0.531 & \cellcolor{myorange}41.129 \\
3DSR (ours) & \cellcolor{myred}24.212 & \cellcolor{myred}0.754 & \cellcolor{myred}0.181 & \cellcolor{myorange}4.632 & \cellcolor{myred}0.516 & \cellcolor{myred}20.731 \\
\bottomrule
\end{tabular}
}
\label{tab:LLFF_quant}
\end{table}

\begin{table}[t]
\centering
\scriptsize
\caption{MipNeRF360 $\times8$ downsampled and $\times4$ upsampled.}
\vspace{-3.5mm}
\resizebox{\columnwidth}{!}{
\begin{tabular}{l|cccccc}
\toprule
\textbf{Method} & \textbf{PSNR}$\uparrow$ & \textbf{SSIM}$\uparrow$ & \textbf{LPIPS}$\downarrow$ & \textbf{NIQE}$\downarrow$ & \textbf{MEt3R}$\downarrow$ & \textbf{FID}$\downarrow$ \\
\midrule
SuperGaussian~\cite{shen2024supergaussian} & \cellcolor{myorange}25.252 & \cellcolor{myorange}0.725 & \cellcolor{myorange}0.303 & \cellcolor{myred}4.694 & \cellcolor{myyellow}0.675 & \cellcolor{myorange}32.652 \\
StableSR~\cite{wang2024exploiting} & \cellcolor{myyellow}24.313 & \cellcolor{myyellow}0.700 & \cellcolor{myyellow}0.326 & \cellcolor{myyellow}5.177 & \cellcolor{myorange}0.644 & \cellcolor{myyellow}44.174\\
3DSR (ours) & \cellcolor{myred}26.097 & \cellcolor{myred}0.746 & \cellcolor{myred}0.222 & \cellcolor{myorange}5.065 & \cellcolor{myred}0.625 & \cellcolor{myred}22.438 \\
\bottomrule
\end{tabular}
}
\label{tab:mipnerf360_ds_8}
\end{table}

        

\section{Conclusion}
We develop a diffusion-based super-resolution framework for 3D representations, leveraging 3D Gaussian Splatting (3DGS) to encourage multi-view consistency. Unlike existing image and video SR methods, which suffer from view-dependent artifacts or require to fine-tune video models, our approach combines diffusion priors with a 3D-consistent rendering pipeline, ensuring structural coherence across viewpoints.

Extensive experiments on MipNeRF360 and LLFF data demonstrate that our method outperforms state-of-the-art ISR and VSR approaches, achieving higher perceptual quality and better preserving geometry. Qualitative results further validate that our framework leads to fewer artifacts, maintains high-frequency textures, and ensures cross-view consistency in novel view synthesis.


{
    \small
    \bibliographystyle{ieeenat_fullname}
    \bibliography{main}
}

\end{document}